\documentclass[11pt,letterpaper]{article}
\usepackage{aaai17}
\usepackage{times}
\usepackage{latexsym}
\usepackage{amsmath}
\usepackage{graphicx}
\usepackage{helvet}
\newcommand{\squeezeup}{\vspace{-1.0mm}}

\title{Discovering Conversational Dependencies between Messages in Dialogs}

\author{Wenchao Du \and Pascal Poupart \\ David R. Cheriton School of Computer Science \\  University of Waterloo \\ \{w8du,ppoupart\}@uwaterloo.ca
         \And 
         Wei Xu \\ Department of Computer Science and Engineering \\ The Ohio State University \\ xu.1265@osu.edu}

\date{}

\begin{document}

\maketitle

\begin{abstract}
We investigate the task of inferring conversational dependencies between messages in one-on-one online chat, which has become one of the most popular forms of customer service. We propose a novel probabilistic classifier that leverages conversational, lexical and semantic information. The approach is evaluated empirically on a set of customer service chat logs from a Chinese e-commerce website.  It outperforms heuristic baselines.  
\end{abstract}

\section{Introduction}




Exposing conversational structure \cite{shen2006thread,elsner2010disentangling} is a key step towards organizing the information in dialogues and is very useful for many applications, such as automatic response generation~\cite{ritter2010unsupervised,sordoni2015neural}
and discourse parsing~\cite{afantenos-EtAl:2015:EMNLP}. There has been a significant rise of interest in conversational response generation using statistical and neural machine translation. These approaches typically require a large number of message-response pairs or context-message-response triples as training data, and such data is usually obtained from human annotations. What remains a challenge is identifying coherent threads of discussion in conversations automatically, which is the goal of this paper. For question answering, finding the relevant context of questions through dependency modelling can help choosing the answer; for text generation, dependency modelling provides 
an effective way to annotate the suitable context-message-response triples for training. 

\begin{figure}[t!]
  \centering
  \includegraphics[width=2in]{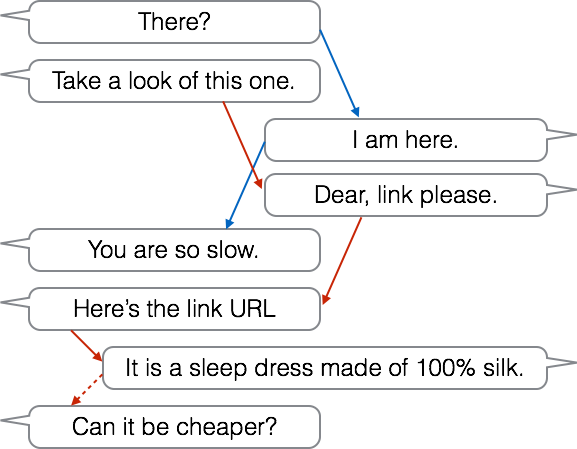}
  \caption{Example of the conversational structure between a customer and a customer service representative (solid arrow is a sure link; dotted arrow is a loose connection)}
  \label{fig:examplechat}
  \vspace{-5mm}
\end{figure}

In online text messaging, one party may send two questions successively, and the other party may answer these questions in any order. In another scenario, one may send a message that does not respond to any of the other party's messages, but elaborate on oneself. These situations complicate the understanding of conversations. Fig.~\ref{fig:examplechat} illustrates
a typical online chat where the correspondence between utterances is crucial for the understanding of conversations.  When annotating the chats, each message is linked to the most relevant message and the candidates comprise of its previous messages and itself. As a result, each chat admits a structure of 1-regular directed graph, which has no cycle except self loops. The fact that the annotation structure is a forest and the number of trees in the forest is not known beforehand eliminates the possibility of applying standard decoding algorithms for discourse parsing such as maximum spanning tree and min-cut.
\squeezeup




\section{Data}
We use the customer service logs from a Chinese e-commerce website. Customers mostly ask about products, promotions, delivery, and sometimes make bargains with agents. Customers also ask for refund after products are delivered. It is not uncommon that text messages from customers are ungrammatical and have spelling errors, which makes using state-of-the-art tagging and dependency parsing tools difficult. Our dataset comprises 9000 chats of 5 to 60 utterances each. We randomly annotated 800 chats of 10 to 35 utterances with an exchange ratio (percentage of consecutive messages by different speakers) between 0.4 and 0.6. We preprocessed the text with an open source word segmenter, python jieba, and replaced non-character lexicons and rare words by their type (e.g., links, emoticons, and geographical names). We only used the most frequent 5,400 words in our training.  Annotation is carried out by 6 annotators who are native speakers. Each annotator is shown 3-5 example chats for training purpose. Each chat is annotated by 3 different people. Annotators were asked to link each message to a previous one with strongest coherence relation in the form of message-response pair, response-continuation pair, question-context pair, or link to itself (if there is no dependency on previous messages). A response-continuation pair is formed only when it is inappropriate to use the continuing message to answer a question directly. When in doubt, annotators were told to think like an agent and to select the most relevant dependency as if they were trying to respond to customers' messages themselves.

Among the 800 chats annotated, 54.2\% received the same label by all 3 annotators and 94.8\% received at least 2 identical labels. Fleiss' Kappa (degree of agreement in classification over that which would be expected by chance) was 0.482. 
The number of classes for each message was 6 (link to itself or any of the last 5 messages).

\squeezeup
\section{Methods}
For each message we consider the following binary features: 1) identity of speaker, 2) contains question words or question mark, 3) contains answer words, 4) contains URL, 5) contains image, 6) contains emoticon. Let $I[f_i^k=a] =1$ when the $k^{th}$ feature of message $i$ takes value $a$ (and 0 otherwise). We also consider the distance between two messages. Let $I[d_{ij}=m]=1$ when message $i$ is $m$ utterances after message $j$ (and 0 otherwise). We define the probability that message $i$ depends on message $j$ as follows:
\begin{equation}
\begin{split}
    p ( & c_i = j | f,d, \mathbf{\eta}, \mathbf{\tau}, \mathbf{\pi}) \propto \\ & \exp(\sum_{k,l,a, b} I[f_{i}^k = a]I[f_{j}^l = b] I[i \not = j]\eta_{klab} + \\
    & \sum_{m} I[d_{ij}=m]\tau_m + \sum_{k, a} I[i = j] I[f_{i}^k = a]\pi_{ka}) 
\end{split}
\label{eq:dist}
\end{equation}
We train the coefficients $\eta_{klab}$, $\tau_m$ and $\pi_{ka}$ by maximizing the conditional likelihood of identifying the correct links in the labeled chats with L2 regularization. This optimization was done by limited memory BFGS implemented in SciPy .  

We then extend the above approach to take into account the semantic similarity between pairs of messages. We apply Latent Dirichlet Allocation (LDA) on the corpus of chats that infers a distribution $\Phi_i$ over latent topics for each message $i$.  We construct a new feature that measures the cross entropy $Cross(i, j) = \sum_t \Phi_{it} log \Phi_{jt}$ between the topic distributions of messages $i,j$. This feature is added to Eq.~\ref{eq:dist} to obtain a new distribution over message dependencies:
\begin{equation}
\begin{split}
    p ( & c_i = j | f,d, \mathbf{\eta}, \mathbf{\tau}, \mathbf{\pi}) \propto \exp\{ \sum_{m} I[d_{ij}=m]\tau_m + \\
    & \sum_{k,l,a, b} I[f_{i}^k = a]I[f_{j}^l = b] I[i \not = j]\eta_{klab} + \\
    & \sum_{k, a} I[i = j] I[f_{i}^k = a]\pi_{ka}) + w Cross(i, j)\}
\end{split}
\end{equation}
We train the new weight $w$ and the other coefficients by maximizing the conditional likelihood of the correct links.
\squeezeup
\section{Experiments}

We compare with two rule-based baselines. Rule1: Each message is linked to its immediate precedent. Rule2: Each message is linked to its immediate precedent if the precedent is from the customer, otherwise it is linked to itself (i.e., customer/agent to customer, but not customer/agent to agent). 

The table below compares our discriminative learning technique with and without the semantic similarity feature from LDA to the baselines. We report the average probability that each method would have labeled a message in the same way as one annotator (chosen uniformly at random among the 3 annotators) based on 5-fold cross validation.  We also report the F1 measure (weighted average of harmonic mean of precision and recall of each class).  Our discriminative learning technique outperforms the baselines, but the semantic similarity feature based on LDA did not yield a significant improvement.  We also estimated human performance by scoring each annotator against the other two annotators, which yielded $0.677 \pm 0.020$.   We can compute an upper bound on the best performance possible by choosing the label with highest agreement among the annotators for each data point, which yielded an accuracy of $0.830$.

\begin{table}[ht]
\begin{center}
\begin{tabular}{|c|c|c|}
    \hline
     & Accuracy & Average F1 \\
     \hline
    Rule-based Baseline 1 & 0.546 & 0.385 \\
    \hline
    Rule-based Baseline 2 & 0.513 & 0.476 \\
    \hline
    Discriminative & 0.624 & 0.580 \\
    \hline
    Discriminative + LDA & 0.626 & 0.588 \\
    \hline
\end{tabular}
\caption{Evaluation}
\end{center}
\end{table}
\squeezeup
\section{Conclusion}

We investigated how to expose the structure of conversations that do not follow perfect turn taking by identifying dependencies between utterances.  We identified a set of relevant features and showed how to train a simple probabilistic model that can infer links.  We obtained encouraging results, but clearly there is still room for improvement.  In particular, we  believe that further improvements based on semantic similarity should be possible and to that effect we are exploring a variant of LDA that will take into account relationships between messages. We will consider the correlation between features.  We also plan to refine the definition of dependencies in order to obtain more consistent annotations which will help to improve the accuracy of the classifiers trained based on those annotations.



\bibliography{emnlp2016}
\bibliographystyle{aaai}

\end{document}